\documentclass[twocolumn,10pt]{asme2ej}

\usepackage{epsfig} 
\makeatletter
\let\NAT@parse\undefined
\makeatother

\usepackage[dvipsnames]{xcolor}

\newcommand*\linkcolours{ForestGreen}
\usepackage{bm}
\usepackage{fixmath}
\usepackage{times}
\usepackage{graphicx}
\usepackage{algorithmic}
\usepackage{algorithm2e}
\usepackage{amssymb}
\usepackage{gensymb}
\usepackage{hyperref}
\usepackage{amsmath}
\usepackage{breakurl}

\usepackage{url,hyperref}
\hypersetup{
colorlinks,
linkcolor=\linkcolours,
citecolor=\linkcolours,
filecolor=\linkcolours,
urlcolor=\linkcolours}

\usepackage[labelfont={bf},font=small]{caption}
\usepackage[none]{hyphenat}

\usepackage{mathtools, cuted}

\usepackage[noadjust, nobreak]{cite}
\def\citepunct{,\,} 

\usepackage{tabularx}
\usepackage{amsmath}

\usepackage{float}

\usepackage{pifont}

\newcolumntype{Y}{>{\centering\arraybackslash}X}

\usepackage[]{placeins}

\usepackage{placeins}
\usepackage{multirow}
\usepackage{multicol}
\usepackage{tikz}

\usepackage[framemethod=tikz]{mdframed}

\usepackage{afterpage}

\usepackage{stfloats}
\usepackage{makecell}
\usepackage{atbegshi}
\newcommand{\handlethispage}{}
\newcommand{\discardpagesfromhere}{\let\handlethispage\AtBeginShipoutDiscard}
\newcommand{\keeppagesfromhere}{\let\handlethispage\relax}
\AtBeginShipout{\handlethispage}

\usepackage{comment}
\newcommand{\newpara}
    {
    \vskip 0.25cm
    }

\title{Comparison of Dynamic and Kinematic Model
Driven Extended Kalman Filters (EKF) for the
Localization of Autonomous Underwater Vehicles}

\author{Sharan Balasubramanian$^{1}$\thanks{These authors contributed equally to this work}, Ayush Rajput$^{1}$\footnotemark[1]~~\thanks{Corresponding author}, Rodra W. Hascaryo$^{2}$, \\ \textbf{Chirag Rastogi}$^{1}$, and \textbf{William R. Norris}$^{1}$ 
   \affiliation{
	$^{1}$ Department of Industrial and Enterprise Systems Engineering\\
	$^{2}$ Department of Aerospace Engineering\\
	University of Illinois at Urbana-Champaign\\
	Urbana, Illinois 61801\\
    Email: [sharanb2, arajput3, hascary2, chiragr2, wrnorris]@illinois.edu
    }
}
    
\begin{document}

\maketitle    

\begin{abstract}
\noindent \textit{Autonomous Underwater Vehicles (AUVs) and Remotely Operated Vehicles (ROVs) are used for a wide variety of missions related to exploration and scientific research. Successful navigation by these systems requires a good localization system. Kalman filter based localization techniques have been prevalent since the early 1960s and extensive research has been carried out using them, both in development and in design. It has been found that the use of a dynamic model (instead of a kinematic model) in the Kalman filter can lead to more accurate predictions, as the dynamic model takes the forces acting on the AUV into account. Presented in this paper is a motion-predictive extended Kalman filter (EKF) for AUVs using a simplified dynamic model. The dynamic model is derived first and  then it was simplified for a RexROV, a type of submarine vehicle used in simple underwater exploration, inspection of subsea structures, pipelines and shipwrecks.  The  filter  was implemented with a simulated vehicle in an open-source marine vehicle  simulator called \textit{UUV Simulator} and the results were compared with the ground truth. The results show good prediction accuracy for the dynamic filter, though improvements are needed before the EKF can be used on real time. Some perspective and discussion on practical implementation is presented to show the next steps needed for this concept.}
\end{abstract}

\section{Introduction}


Localization is one of the key pillars of autonomous navigation. A major component of localization is the ability of the system to estimate the state of the vehicle in order to help locate it. Bayesian estimators are most commonly used for this, the most common strategy using Bayesian estimators being the Kalman filter (KF). The KF assumes the states to be Gaussian random variables and uses least squares minimization to estimate final state. For under-water localization, the Extended Kalman filter (EKF) has been widely used~\cite{7003060, 6146033, s20174710}. An EKF linearizes a non-linear dynamic model to find a local approximation. 

Sampling-based strategies are also popular choices for state estimation. The Unscented Kalman filter (UKF) and particle filter (PF) are two common choices. The UKF is an improvement over the EKF for some applications, using sigma points to estimate the state parameters. It uses a deterministic sampling strategy but assumes underlying the state to be a Gaussian random variable. Comparing the performance of EKF with UKF, \cite{7271681} observed UKF to be better while \cite{6510082} has observed EKF to be better, so the appropriate one to use depends on the inputs and application. When underlying state estimation is not a Gaussian random variable, particle filter (PF) is a good choice. The particle filter represents a non-Gaussian state distribution using weighted samples, and uses a random sampling strategy. PF has been successfully implemented to estimate states for underwater vehicles in several studies, including~\cite{5152014, 6263573, 6853000}. In comparison to a Kalman filter, localization is observed to be more robust when PF is used under sensor noise, as observed by~\cite{6463013}.

In state estimation, the plant (hardware) is most commonly represented by a purely kinematic model (i.e., model describing vehicle motion without considering force and moments~\cite{6510082, 1545230, 4383217}). However kinematic models fail to capture highly non-linear behaviour often observed in underwater systems. Approaches based on dynamic models have been implemented successfully \cite{spain, 5773655, 7271681, 5773655} to get more accurate state estimates. The study described in Ref.~\cite{5773655} implemented a 3 DoF dynamic model along with INS and DVL sensor data to estimate state of vehicle.



\begin{figure}[h]
\centering
\includegraphics[width=8cm,height=4cm,keepaspectratio]{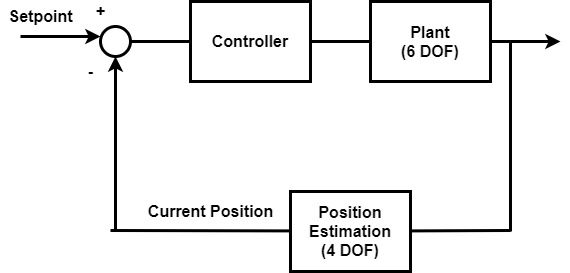}
\caption{Block Diagram}
    \label{fig:block_diag}
\centering
\end{figure}

\begin{figure}[h]
\centering
\includegraphics{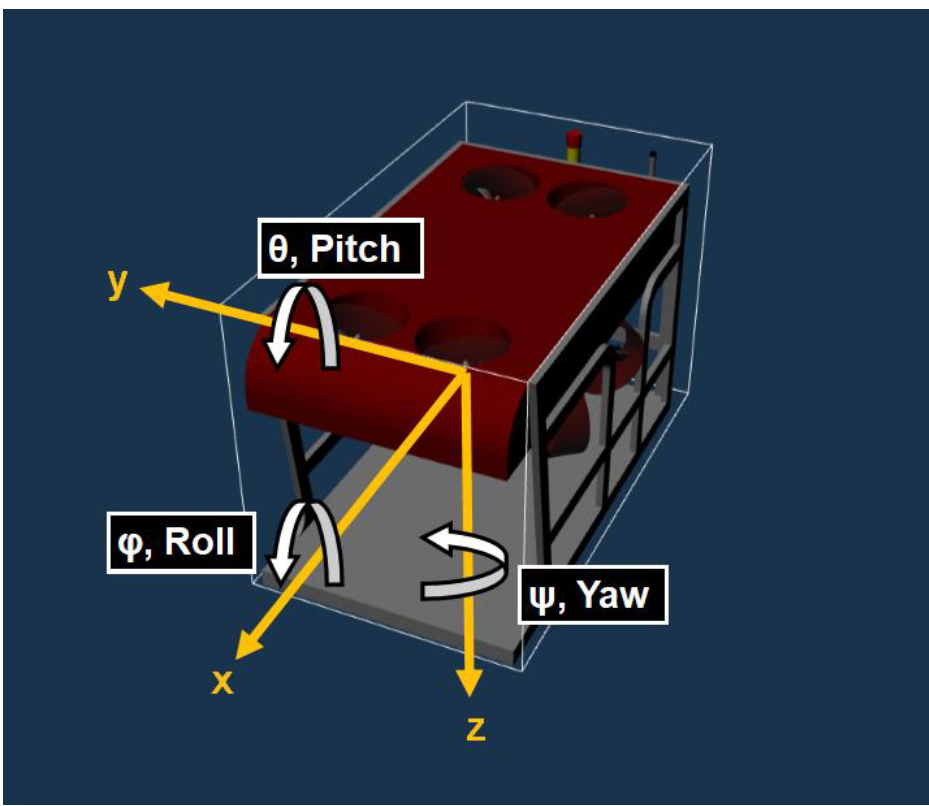}
\caption{BODY axis convention}
    \label{fig:BODY}
\centering
\end{figure}

To explore this further, this article derives a rigorous dynamic model of an underwater vehicle (UWV) to be used with an extended Kalman filter. This model considers several forces acting on the UWV and is simplified to be able to work on a RexROV vehicle within the UUVSim~\cite{uuvsim} simulation environment. This simplification was done to reduce the number of terms in the dynamic model, since some of the vehicle degrees-of-freedom (DoF) are not used. Removing these terms reduces the computational cost while only eliminating terms which had little to no effect on the vehicle behavior. The RexROV vehicle, as shown in Fig.~\ref{fig:BODY}, has 4 lateral thrusters and 4 longitudinal thrusters to govern its motion. The RexROV model in UUVSim consists of the dynamic model with GPS, Doppler Velocity Logs (DVL) and IMU sensors. The block diagram in Fig.~\ref{fig:block_diag} summarizes the proposed digital experiments. In this simulation, a vehicle with 6 DoF is set for waypoint navigation. It is localized using a 4 DoF dynamic model with an EKV. A pure pursuit controller is used to generate control commands.

The article is divided into several sections, beginning with a detailed description of the methodology (Section 2), the definition of the coordinate system and conventions (Section 3), the derivation (Section 4) and simplification (Section 5) of the dynamic model, a discussion of the IMU acceleration (Section 6), the EKF algorithm (Section 7), a description of the controller (Section 8), some simulated experiments to test the performance (Section 9), a discussion of practical implementation, leading to physical real-world experiments (Section 10), and finally some conclusions and future work directions (Section 11).

\section{Methodology}

Beginning with the seminal work by Fossen~\cite{Fossen_book}, kinematic and dynamic models, which have been observed to be the \textit{de facto} standard used in marine vehicle modeling, are laid out in Refs.~~\cite{uuvsim, joint, Berg, eleven} and then simplified for a particular vehicle before being applied to an EKF and run in a computer simulation. The work cited above use a 6 DoF dynamic model for the application, although Berg~\cite{Berg} noted that the ROV effectively has 4 DoF~\cite{Berg}. This research aims to investigate the feasibility of a 4 DoF model for motion prediction by incorporating it into an EKF. Two DoF were eliminated due to the inherent stability of the vehicle which  helped in reducing the 6x6 state matrix to a pseudo 4x4 state matrix thereby reducing computational cost.The 4 DOF model is proposed in Ref.~\cite{2000} and was successfully used by Ref.~\cite{kim15} in control development. Since motion predictions are being augmented with sensor readings, it is expected from this approach to work effectively with the problem proposed in this paper. A Robotic Operating System (ROS) Gazebo-based open-source marine vehicle simulator was used in  this  research~\cite{uuvsim}. The  simulator incorporates the dynamic model by  Fossen~\cite{uuvsim} with a vehicle model  based on parameters derived  from  the  works by  Berg~\cite{Berg}.

\vspace{-10pt}

\section{Coordinate Systems and Convention}
\newpara
Before derivation of the dynamic model, it is important to define and describe the coordinate system that should be used for it. The simulation consists of two bodies - the world or environment and the underwater vehicle, each with its own coordinate system. The world uses a fixed-frame North-East-Down (NED) and the vehicle uses a fixed-frame BODY~\cite{joint, Berg, Fossen_book} system. The BODY frame is based on the Society of Naval Architects and Marine Engineers (SNAME) 1950 convention~\cite{Fossen_book}; the BODY axis convention for RexROV is as shown in Fig.~\ref{fig:BODY}. The NED frame is used for the vehicle's position, pose and the environment related dynamic effects such as buoyancy and weight~\cite{joint}, while the BODY frame is used to describe the vehicle's linear and angular velocities~\cite{Berg}. Table~\ref{table:SNAME Convention} shows the naming convention followed.

\begin{table}[htb]
\centering
\caption{1950 SNAME convention for marine vehicles}\label{table:SNAME Convention}
\begin{tabular}{|l|c|c|c|c|}
\hline
\textbf{Term}&\textbf{Pose}&\textbf{Velocity}&\textbf{Acceleration}&\textbf{Force}  \\
\hline
Surge&x&u&$\dot{u}$&X\\
\hline
Sway&y&v&$\dot{v}$&Y\\
\hline
Heave&z&w&$\dot{w}$&Z\\
\hline
Roll&$\phi$&p&$\dot{p}$&K\\
\hline
Pitch&$\theta$&q&$\dot{q}$&M\\
\hline
Yaw&$\psi$&r&$\dot{r}$&N\\
\hline
\end{tabular}
\end{table}

\noindent Clear transformation between the two coordinate systems is crucial to the dynamic model. Considering~\eqref{eq:1}, the transformation from BODY to NED coordinates is:
\begin{equation}
    \dot{\eta}_{b/n}^{n} = J_{b}^{n}(\Theta)v_{b/n}^{b}
    \label{eq:4}
\end{equation}
where subscript \textbf{\textit{b/n}} means frame \textbf{\textit{b}} in motion with respect to frame \textbf{\textit{n}}. The Superscript denotes the frame in which matrix is defined. Matrix J is a $6\times6$ transformation matrix given:
\begin{equation}
    J_{b}^{n}(\Theta)=\begin{bmatrix}R_{b}^{n}(\Theta) & 0_{3x3} \\0_{3x3} & T_{b}^{n}(\Theta)
    \end{bmatrix}
    \label{eq:5}
\end{equation}

\noindent where $R_{b}^{n}(\Theta)$ and $T_{b}^{n}(\Theta)$ are the following $3\times3$ rotation matrices for linear and angular velocities with a condition $\cos{\it{A}}\neq$ 0 thus \textit{A} $\neq$ 90\degree or 270\degree~\cite{Berg}.

\begin{equation}
    R_{b}^{n}(\Theta)=
    \begin{bmatrix}
    c{\psi}c{\theta} & s{\phi}s{\theta}c{\psi}-s{\psi}c{\phi} & s{\psi}s{\phi}+c{\psi}c{\phi}s{\theta}\\
    s{\psi}c{\theta} & c{\psi}c{\phi}+s{\phi}s{\theta}s{\psi} & s{\theta}s{\psi}c{\phi}-s{\phi}c{\psi}\\
    -s{\theta} & s{\phi}c{\theta} & c{\theta}c{\psi}
    \end{bmatrix}
    \label{eq:6}
\end{equation}

\begin{equation}
    T_{b}^{n}(\Theta)=
    \begin{bmatrix}
    1 & s{\phi}t{\theta} & c(\phi)t{\theta}\\
    0 & c{\phi} & -s{\phi}\\
    0 & s{\phi}c{\theta} & c{\phi}c{\theta}
    \end{bmatrix}
    \label{eq:7}
\end{equation}

\noindent For simplicity, cosine, sine, and tangent of angle \textit{A} has been shown as c\textit{A}, s\textit{A}, and t\textit{A} respectively above.
\section{Dynamic Model derivation}

The forces acting on an underwater vehicle can be divided into five categories: Kinetic forces, hydrodynamics, hydrostatics, actuator forces and disturbances. The 6 DoF dynamics of an underwater vehicle can be represented with the following two vector equations~\cite{Fossen_phd}:
\begin{equation}
   \dot{\eta} =J(\eta)\mathit{v}\label{eq:1}
\end{equation}

\begin{equation}
    M\dot{\mathit{v}} + C(\mathit{v})\mathit{v} + D(\mathit{v})\mathit{v} + g(\eta) + g_0 = \tau + \tau_{wind} + \tau_{wave}\label{eq:2}
\end{equation}


\noindent The terms used in the equation are as listed below:
\begin{table}[htbp]
\begin{tabular}{l l l}
$\eta$ & - & Vehicle position and pose \\
$\textit{v}$ & - & Vehicle velocity (in BODY frame)\\
$J$ & - & \makecell[l]{Transformation matrix}\\
$M$ & - & inertial matrix\\
$C\textit{(v)}$ & - & Coriolis forces\\
$D\textit{(v)}$ & - & Damping matrix\\
$g(\eta)$ & - & Buoyancy force \\
$\tau$ & - & Actuator force \\
$\tau_{wind}$ & - & Effect of wind \\
$\tau_{wave}$ & - & Effect of water waves \\
\end{tabular}
\end{table} 
\vspace{0.25cm}

\noindent The dynamic model consists of the terms and coefficients listed in the table~\ref{table:Parameter}.

\begin{table}[H]
    \centering
    \caption{List of Parameter and Coefficient Symbols}
    \label{table:Parameter}
    \begin{tabular}{|l| m{145pt} |}
\hline
\textbf{Coefficient} &\textbf{Explanation}  \\
\hline
\it{m} & Mass of the vehicle \\
\hline
\it{W} & Weight of the vehicle \\
\hline
\it{B} & Buoyancy force \\
\hline
\it{g} & Acceleration due to gravity \\
\hline
\it{$\rho$} & Density of water \\
\hline
\it{I} & Moment of inertia \\
\hline
$I_x,I_y,I_z$ & Moment arm of thrusters from center of gravity \\
\hline
$X_{\dot{u}},Y_{\dot{v}},Z_{\dot{w}}$ & Mass increase due to translation \\
\hline
$K_{\dot{p}},M_{\dot{q}},N_{\dot{r}}$ & Inertia increase due to rotation \\
\hline
$X_{{u}},Y_{{v}},Z_{{w}}$ & Linear damping coefficients for translation\\
\hline
$X_{{u|u|}},Y_{{v|v|}},Z_{{w|w|}}$ & Quadratic damping coefficients for translation\\
\hline
$K_{{p}},M_{{q}},N_{{r}}$ & Linear damping coefficients for rotation\\
\hline
$K_{{p|p|}},M_{{q|q|}},N_{{r|r|}}$ & Quadratic damping coefficients for rotation \\
\hline
\end{tabular}
\end{table}

The forces acting on an underwater vehicle are modeled separately at first and then substituted in equation~\eqref{eq:2} to derive the dynamic model.
The vehicle is considered as a rigid body since it's not expected to undergo any deformation at any operating depth. Thus the rigid body kinetics ($\tau_{RB}$) can be expressed as a sum of the inertial ($M_{RB}\dot{\mathit{v}})$ and Coriolis effects ($C_{RB}(\mathit{v})\mathit{v})$ as shown in equation~\eqref{eq:3}.
\begin{equation}
    \tau_{RB}=M_{RB}\dot{\mathit{v}} + C_{RB}(\mathit{v})\mathit{v}
    \label{eq:3}
\end{equation}

\noindent For a 6 DoF model, where the vehicle center of origin and center of gravity coincide, $M_{RB}$ can be represented as in~\eqref{eq:8}.

\begin{equation}
M_{RB}=
  \begin{bmatrix}
    m & 0 & 0 & 0 & 0 & 0 \\
    0 & m & 0 & 0 & 0 & 0 \\
    0 & 0 & m & 0 & 0 & 0 \\
    0 & 0 & 0 & I_{xx} & 0 & 0 \\
    0 & 0 & 0 & 0 & I_{yy} & 0 \\
    0 & 0 & 0 & 0 & 0 & I_{zz} \\
  \end{bmatrix}
  \label{eq:8}
\end{equation}
    
\noindent The Coriolis effect($C_{RB}(\mathit{v})$) is due to the rotation of the environment in which the AUV is operating and hence it can be represented as in~\eqref{eq:9}.

\begin{equation}
C_{RB}(\mathit{v})=
  \begin{bmatrix}
    mS(\mathit{v_2}) & -mS(\mathit{v_2})S(\mathit{r_g^b}) \\
    mS(\mathit{r_g^b})S(\mathit{v_2}) & -S(I_b \mathit{v_2}) \\
  \end{bmatrix}
  \label{eq:9}
\end{equation}

\noindent where $r_g^b$ is the position of the Center of Gravity (CG) matrix. The vector $\mathit{v_2}$ (velocity vector) is the velocity vector and $I_b$ is the moment of inertia matrix and $S$ represents the skew-symmetric matrix. Fossen~\cite{Fossen_book, Fossen_phd} showed that multiplying the rigid body Coriolis matrix above with the velocity vector yield the Coriolis effect as shown in~\eqref{eq:10}.

\begin{equation}
C_{RB}(\mathit{v})\mathit{v}= 
  \begin{bmatrix}
m(qw-rv) \\
m(ru-pw) \\
m(pv-qu) \\
qr(I_{zz}-I_{yy}) \\
rp(I_{xx}-I_{zz}) \\
qp(I_{yy}-I_{xx})\\
  \end{bmatrix}
  \label{eq:10}
\end{equation}

\noindent The hydrodynamic force ($\tau_{hydrodynamics}$) includes the hydrodynamic drag ($D(\mathit{v})\mathit{v}$), the Coriolis-centripetal effects ($C_A(\mathit{v})\mathit{v}$) and the inertial effects ($M_A\dot{\mathit{v}}$) as shown in~\eqref{eq:11}~\cite{Fossen_book}.

\begin{equation}
    \tau_{hydrodynamics}= -D(\mathit{v})\mathit{v}-C_A(\mathit{v})\mathit{v}-M_A\dot{\mathit{v}}
    \label{eq:11}
\end{equation}

\noindent The individual components of $\bm{\tau_{hydrodynamics}}$ are simplified from Fossen equations using the low speed assumption~\cite{Fossen_book}: 
\begin{equation}
M_A=-
  \begin{bmatrix}
    X_{\dot{u}} & 0 & 0 & 0 & 0 & 0 \\
    0 & Y_{\dot{v}} & 0 & 0 & 0 & 0 \\
    0 & 0 & Z_{\dot{u}} & 0 & 0 & 0 \\
    0 & 0 & 0 & K_{\dot{p}} & 0 & 0 \\
    0 & 0 & 0 & 0 & M_{\dot{q}} & 0 \\
    0 & 0 & 0 & 0 & 0 & N_{\dot{r}} \\
  \end{bmatrix}
  \label{eq:12}
\end{equation}

\begin{equation}
C_A(\mathit{v})=
\setlength\arraycolsep{0.01pt}
  \begin{bmatrix}
    0 & 0 & 0 & 0 & -Z_{\dot{w}} w & Y_{\dot{v}}v\\
    0 & 0 & 0 & Z_{\dot{w}}w & 0 & -X_{\dot{u}}u \\
    0 & 0 & 0 & -Y_{\dot{v}}v & X_{\dot{u}}u & 0 \\
    0 & -Z_{\dot{w}}w & Y_{\dot{v}}v & 0 & -N_{\dot{r}}r & M_{\dot{q}}q \\
     Z_{\dot{w}}w & 0 & -X_{\dot{u}}u & N_{\dot{r}}r & 0 & -K_{\dot{p}}p \\
    -Y_{\dot{v}}v & X_{\dot{u}}u & 0  & -M_{\dot{q}}q  & K_{\dot{p}}p & 0 \\
  \end{bmatrix}
  \label{eq:13}
\end{equation}

\noindent Multiplying~\eqref{eq:13} with the velocity vector:

\begin{equation}
C_A(\mathit{v})\mathit{v}=
\setlength\arraycolsep{0.01pt}
  \begin{bmatrix}
    Y_{\dot{v}}vr -  Z_{\dot{w}}wq \\
     Z_{\dot{w}}wp - X_{\dot{u}}ur\\
     X_{\dot{u}}uq - Y_{\dot{v}}vp\\
     (Y_{\dot{v}}-Z_{\dot{w}})vw +(M_{\dot{q}}-N_{\dot{r}})qr \\
     (Z_{\dot{w}}-X_{\dot{u}})uw +(N_{\dot{r}}-K_{\dot{p}})pr \\
     (X_{\dot{u}}-Y_{\dot{v}})uv+(K_{\dot{p}}-M_{\dot{q}})pq 
  \end{bmatrix}
  \label{eq:14}
\end{equation}
The drag force consists of two components namely linear and quadratic drag forces as shown in equation~\eqref{eq:15}~\cite{Fossen_book}. 
\begin{equation}
    D(\mathit{v})= D_{linear}+D_{quadratic}
    \label{eq:15}
\end{equation}
where $D_{linear}$ is a diagonal matrix with all the linear drag components and $D_{quadratic}$ is a diagonal matrix with all the quadratic components. Thus summing the linear and quadratic drag terms and multiplying them with the velocity components $D(\mathit{v})\mathit{v}$ is found as shown in~\eqref{eq:16}. 

\begin{equation}
  D(\mathit{v})=- 
  \begin{bmatrix}
  (X_u +X_{u|u|}|u|)u\\
  (Y_v +Y_{v|v|}|v|)v\\
  (Z_w +Z_{w|w|}|w|)w\\
  (K_p +K_{p|p|}|p|)p\\
  (M_q +M_{q|q|}|q|)q\\
  (N_r +N_{r|r|}|r|)r
  \end{bmatrix}
  \label{eq:16}
\end{equation}

The hydrostatic forces consist of two forces: buoyancy and gravitational force(weight). It is assumed that the Center of Buoyancy (CB) and Center of Gravity (CG) are on the z-axis and  $z_{B}$ is the distance of the CB to the CG for simplicity as the assumption holds for most of the cases. Buoyancy is the upward force caused when immersing a body into a liquid and is a factor of the volume of the liquid displaced by the body. In the case of a fully submerged vehicle (AUV), the volume of liquid displaced is equal to the total volume of the vehicle. Hence buoyancy (B) is as shown in~\eqref{eq:17}. The weight (W) of the vehicle is equal to the mass times acceleration due to gravity as shown in~\eqref{eq:18}.
\begin{equation}
\label{eq:17}
    B = {\rho}gV 
\end{equation}
\begin{equation}
    W = mg \label{eq:18}
\end{equation}

\noindent In the case of an AUV, \textit{${\rho}$} is the density of water and \textit{V} is the volume of the vehicle. The sum of the two forces gives the hydrostatic force which in vector form, is written as (\ref{eq:19}).
\begin{equation}
  g(\eta)=
  \begin{bmatrix}
  (W-B)\sin{\theta}\\
  -(W-B)\cos{\theta}\sin{\phi}\\
  -(W-B)\cos{\theta}\cos{\phi}\\
  -z_{B}B\cos{\theta}\sin{\phi}\\
  -z_{B}B\sin{\theta}\\
  0
  \end{bmatrix}
  \label{eq:19}
\end{equation}

The final component of the dynamic model involves calculating the actuator forces. Deep underwater, the effect of wind and ocean currents are minimal and so $\tau_{wind}$ and $\tau_{wave}$ are equivalent to zero. The thruster forces are vehicle specific since different vehicles have different specifications and orientations for their thrusters. In general, the force and the torque contribution of each thruster in the 6 DoF is represented as shown in equation (\ref{eq:20}).
\begin{equation}
  T_{i}= 
  \begin{bmatrix}
  T_{x}\\
  T_{y}\\
  T_{z}\\
  T_{\phi}\\
  T_{\theta}\\
  T_{\psi}
  \end{bmatrix}
  \label{eq:20}
\end{equation}

There are both empirical and theoretical methods to measure these thruster forces~\cite{joint, Berg, eleven}. Finally substituting the above derived formulas into equation (\ref{eq:2}) and decomposing them into respective components would yield the dynamic model as shown in equations~\eqref{eq:21} to~\eqref{eq:26}.

\begin{equation}
\begin{split}
        \dot{u}=  \bigg(\frac{1}{m-X_{\dot{u}}}\bigg) \big((X_{u}+X_{u|u|}|u|)u + (B-W)\sin{\theta}\\ 
         \cap +m(rv-qw)  - Y_{\dot{v}}rv +  Z_{\dot{w}}qw + T_x \big)
\end{split}
\label{eq:21}
\end{equation}

\begin{equation}
\begin{split}
        \dot{v}=  \bigg(\frac{1}{m-Y_{\dot{v}}}\bigg) \big((Y_{v}+Y_{v|v|}|v|)v - (B-W)\sin{\phi}\cos{\theta}\\ 
         \cap +m(pw-ru)  + X_{\dot{u}}ru - Z_{\dot{w}}pw + T_y \big)
\end{split}
\label{eq:22}
\end{equation}

\begin{equation}
\begin{split}
        \dot{w}=  \bigg(\frac{1}{m-_Z{\dot{w}}}\bigg) \big((Z_{w}+Z_{w|w|}|w|)w - (B-W)\cos{\theta}\cos{\phi}\\ 
         \cap +m(qu-pv)  - X_{\dot{u}}qu + Y_{\dot{v}}pv + T_z \big)
\end{split}
\label{eq:23}
\end{equation}

\begin{equation}
\begin{split}
        \dot{p}=  \bigg(\frac{1}{I_{x}-K_{\dot{p}}}\bigg) \big((K_{p}+K_{p|p|}|p|)p - (M_{\dot{q}}-N_{\dot{r}})qr\\ 
         \cap +(I_{yy}-I_{zz})qr + (Z_{\dot{w}}-Y_{\dot{v}})vw \\ 
         \cap+ Bz_{B}\cos{\theta}\sin{\phi} + T_{\theta} \big)
\end{split}
\label{eq:24}
\end{equation}

\begin{equation}
\begin{split}
        \dot{q}=  \bigg(\frac{1}{I_{y}-M_{\dot{q}}}\bigg) \big((M_{q}+M_{q|q|}|q|)q - (N_{\dot{r}}-K_{\dot{p}})pr\\ 
         \cap +(I_{zz}-I_{xx})pr + (X_{\dot{u}}-Z_{\dot{w}})uw\\ 
         \cap + Bz_{B}\sin{\theta}  + T_{\phi} \big)
\end{split}
\label{eq:25}
\end{equation}

\begin{equation}
\begin{split}
        \dot{r}=  \bigg(\frac{1}{I_{z}-N_{\dot{r}}}\bigg) \big((N_{r}+N_{r|r|}|r|)r - (K_{\dot{p}}-M_{\dot{q}})pq\\ 
         \cap +(I_{xx}-I_{yy})pq + (Y_{\dot{v}}-X_{\dot{u}})uv  + T_{\psi} \big)
\end{split}
\label{eq:26}
\end{equation}


\section{Reduced Dynamic Model for RexROV}
\begin{figure}[h]
\centering
\includegraphics{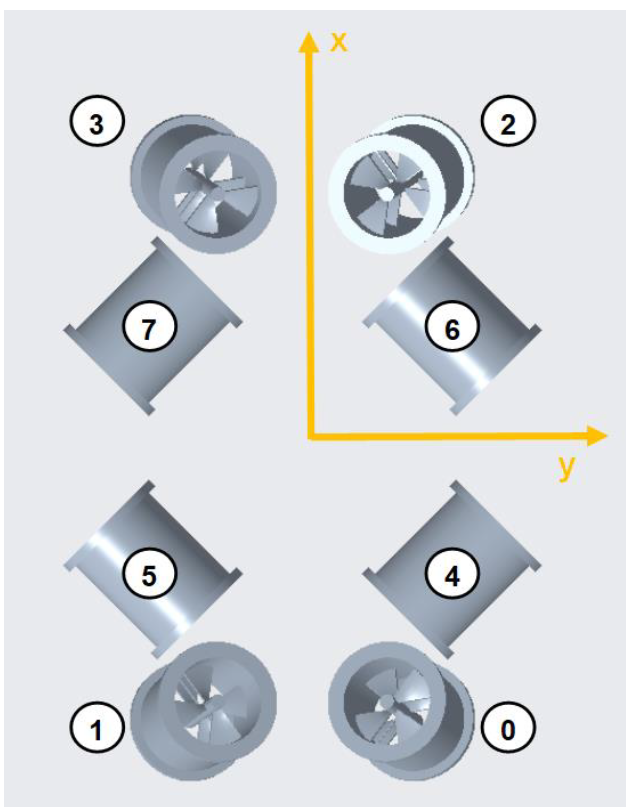}
\caption{RexROV's thrusters orientation from UUVSim\cite{rodra}}
    \label{fig:Thrusters}
\centering
\end{figure}

The RexROV submarine is the AUV used in this research and it relies on its thrusters to orient and move itself. The position and orientation of the thrusters are as shown in Fig.~\ref{fig:Thrusters}.

Through the inherent configuration of the RexROV, the submarine is stable with no pitch and roll due to its structure. From an operations perspective, there is also no need for the RexROV or similar vehicles to possess pitch and roll capabilities. Thus 2 DoF can be eliminated and the 6 DoF model can be reduced to a pseudo 4 DoF model. This means that $\phi$ and $\theta$ are always 0. Likewise, their corresponding velocities, p and q, and accelerations, $\dot{p}$ and $\dot{q}$, components are always 0. Canceling those terms for the other four linear and angular accelerations, the dynamic model \footnote[1]{ It means equation continues $\bm{\cap}$ } becomes:

\begin{equation}
\begin{split}
        \dot{u}=  \bigg(\frac{1}{m-X_{\dot{u}}}\bigg) \big((X_{u}+X_{u|u|}|u|)u \\ 
         \cap +mrv  - Y_{\dot{v}}rv +  T_x \big)
\end{split}
\label{eq:27}
\end{equation}

\begin{equation}
\begin{split}
        \dot{v}=  \bigg(\frac{1}{m-Y_{\dot{v}}}\bigg) \big((Y_{v}+Y_{v|v|}|v|)v \\ 
         \cap -mru + X_{\dot{u}}ru + T_y \big)
\end{split}
\label{eq:28}
\end{equation}

\begin{equation}
\begin{split}
        \dot{w}=  \bigg(\frac{1}{m-_Z{\dot{w}}}\bigg) \big((Z_{w}+Z_{w|w|}|w|)w
        \\ 
         \cap- (B-W) + T_z \big)
\end{split}
\label{eq:29}
\end{equation}

\begin{equation}
\begin{split}
        \dot{r}=  \bigg(\frac{1}{I_{z}-N_{\dot{r}}}\bigg) \big((N_{r}+N_{r|r|}|r|)r \\ 
         \cap+ (Y_{\dot{v}}-X_{\dot{u}})uv  + T_{\psi} \big)
\end{split}
\label{eq:30}
\end{equation}


A Creo based CAD model was developed based on the information from the UUVSim website~\cite{uuvsim}. The tables~\ref{table:Thruster Positions} and~\ref{table:Thruster orientation} show the Thruster ID and their respective location from the CG with their orientations.

\begin{table}[!htbp]
    \caption{RexROV Thruster Positions}
    \centering
    \begin{tabular}{|c|c|c|c|}
    \hline
    Thruster  & \multicolumn{3}{|c|}{Location w.r.t. CG(m)}\\ 
    \cline{2-4}
    ID&lx&ly&lz\\
    \hline
    0&-0.890895&0.334385&-0.528822\\
    \hline
    1&-0.890895&-0.334385&-0.528822\\
    \hline
    2&0.890895&0.334385&-0.528822\\
    \hline
    3&0.890895&-0.334385&-0.528822\\
    \hline
    4&-0.412125&0.505415&-0.129\\
    \hline
    5&-0.412125&-0.505415&-0.129\\
    \hline
    6&0.412125&0.505415&-0.129\\
    \hline
    7&0.412125&-0.505415&-0.129\\
    \hline
    \end{tabular}

    \label{table:Thruster Positions}
\end{table}

\begin{table}[!htbp]
    \caption{RexROV Thruster Orientation}
    \centering
    \begin{tabular}{|c|c|c|c|}
    \hline
    {Thruster}  & \multicolumn{3}{|c|}{Orientation(deg)} \\
    \cline{2-4}
    ID&${\phi}$&${\theta}$&${\psi}$\\
    \hline
    0&0&74.53&-53.21\\
    \hline
    1&0&74.53&53.21\\
    \hline
    2&0&105.47&53.21\\
    \hline
    3&0&105.47&-53.21\\
    \hline
    4&0&0&45\\
    \hline
    5&0&0&45\\
    \hline
    6&0&0&135\\
    \hline
    7&0&0&-135\\
    \hline
    \end{tabular}
    \label{table:Thruster orientation}
\end{table}

Based on Fig.~\ref{fig:Thrusters} and tables~\ref{table:Thruster Positions} and~\ref{table:Thruster orientation}, the individual thruster forces $T_{x}$, $T_{y}$, $T_{z}$ and $T_{\psi}$ can be derived by resolving them based on their respective orientations as shown in~\eqref{eq:31} to~\eqref{eq:34}.

\begin{equation}
T_{x} =  \sum_{i=0}^{3} T_i\cos{\theta_i}\cos{\psi_i} + \sum_{j=4}^{7} T_j\cos{\psi_j} 
\label{eq:31}
\end{equation}

\begin{equation}
T_{y}= \sum_{i=0}^{3} T_i\cos{\theta_i}\cos{\psi_i} + \sum_{j=4}^{7} T_j\sin{\psi_j}
\label{eq:32}
\end{equation}

\begin{equation}
\begin{split}
T_{z}=  \sum_{i=1}^{3} T_i\sin{\psi_i}
\end{split}
\label{eq:33}
\end{equation}

\begin{equation}
\begin{split}
T_{\psi}= \sum_{i=0}^{7} T_i(lx_{i}\sin{\psi_i} - ly_{i}\cos{\psi_i})
\end{split}
\label{eq:34}
\end{equation}

The angle terms \textit{$\theta_A$} and \textit{$\psi_A$} used in the above four equation are the orientation of thruster A with respect to the xy-plane and z-axis respectively. They are taken from Tables~\ref{table:Thruster Positions} and~\ref{table:Thruster orientation}. Tables~\ref{table:Rexrovphysicalpara} and~\ref{table:Rexrovcoeffpara} provide the RexROV parameters and coefficients that are used in the equations~\eqref{eq:27} to~\eqref{eq:30}. These coefficients are specific to RexROV as shown in~\ref{fig:BODY} and are derived from Berg's model~\cite{Berg} and UUVSim~\cite{uuvsim}.

\begin{table}[h]
    \caption{RexROV Physical Parameters}
    \centering
    \begin{tabular}{|l|l|}
    \hline
    $\textbf{Parameter}$ & $\textbf{Value}$  \\
    \hline
    Length(x-axis) & 2.6 m \\
    \hline
    Width(y-axis) & 1.5 m\\
    \hline
    Height(z-axis) & 1.6 m\\
    \hline
    Mass(m) & 1.863 kg\\
    \hline
    Volume(V) & 1.838 $m^{3}$\\
    \hline
    B & 18393.9972 N\\
    \hline
    g & 9.81 $m/s$ \\
    \hline
    $\rho$ & 1000 $kg/m^{3}$\\
    \hline
    $I_{zz}$ & 691.23 $kg m^{2}$\\
    \hline
    \end{tabular}
    \label{table:Rexrovphysicalpara}
\end{table}

\begin{table}[h]
    \caption{RexROV Coefficients}
    \centering
    \begin{tabular}{|l|l|}
    \hline
    $\textbf{Coefficient}$ &$\textbf{Value}$  \\
    \hline
    $X_{\dot{u}}$ & 779.79\\
    \hline
    $Y_{\dot{v}}$ & 1222\\
    \hline
    $Z_{\dot{w}}$ & 3659.9\\
    \hline
    $K_{\dot{p}}$ & 534.9\\
    \hline
    $M_{\dot{q}}$ & 842.69\\
    \hline
    $N_{\dot{r}}$ & 224.32\\
    \hline
    $X_{{u}}$ & -74.82\\
    \hline
    $Y_{{v}}$ & -69.48\\
    \hline
    $Z_{{w}}$ & -782.4\\
    \hline
    $K_{{p}}$ & -268.8\\
    \hline
    $M_{{q}}$ & -309.77\\
    \hline
    $N_{{r}}$ & -105\\
    \hline
    $X_{{u|u|}}$ & -748.22\\
    \hline
    $Y_{{v|v|}}$ & -992.53\\
    \hline
    $Z_{{w|w|}}$ & -1821.01\\
    \hline
    $K_{{p|p|}}$ &-672\\
    \hline
    $M_{{q|q|}}$ &-774.44\\
    \hline
    $N_{{r|r|}}$ &-523.27\\
    \hline
\end{tabular}
    \label{table:Rexrovcoeffpara}
\end{table}
\section{IMU acceleration}
As seen in equations~\eqref{eq:27} to~\eqref{eq:30}, the dynamic model for RexROV involves calculating the acceleration values. The accuracy of the modeling can be verified by comparing it to the acceleration values generated by the IMU sensor on UUVSim~\cite{uuvsim}. The IMU sensor takes into account drifting bias by following a random walk with zero-mean white Gaussian noise as the rate. Its model includes zero-mean Gaussian noise as well.

\begin{figure}[h]
\centering
\includegraphics[scale=0.55]{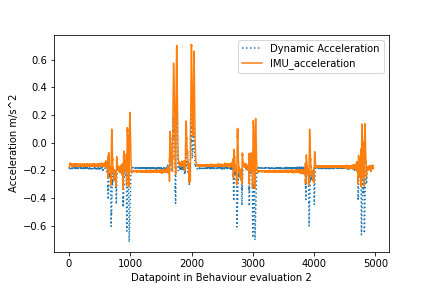}
\caption{Acceleration values predicted by model and IMU in x-axis}
    \label{fig:imu-x}
\centering
\end{figure}

The following Fig.~\ref{fig:bp3} shows the acceleration values generated by the model and the acceleration values given by the IMU sensor in the simulation in the x-axis for behavior evaluation 3 test course~\cite{Norris}. It can be seen that our model performs similarly to the IMU sensor. It is also important to note that the IMU sensor modeled in the simulation has a very high level of accuracy with a variance of 0.004 m/s\textsuperscript{2}. However, in practice, the IMU sensors have a higher variance than the one in simulation. In general, IMUs are prone to drift, which results in error accumulation over time. This results in divergence with actual values in practice while calculating position.

As seen in Fig.~\ref{fig:imu-x}, the dynamic model provides a change in the value where there is a change in the trajectory. The positive acceleration values calculated by the model are consistent with the IMU sensor values. Further research will be conducted to understand the inconsistency on the negative acceleration values calculated by the model. The dynamic model has an edge over the IMU sensor in the fact that it uses velocity and thrust forces for calculation. The thrust forces can be reliably received from the generated look-up table based on the thruster RPM. The velocity, being the first derivative, gives a highly reliable estimate for the position calculation as opposed to acceleration, which is the second derivative.

\section{Extended Kalman Filter Algorithm}
The Kalman Filter~\cite{Kalman} is an iterative mathematical tool used to estimate the state of a system based on a system of equations and measurements. The Kalman filter has three steps, namely, the Initialization, Prediction and Update. For a Kalman filter to work, there are two major conditions. The Kalman filter works with the Gaussian distribution with linear prediction and update equations. These two conditions are interrelated. A linear system fed to a Gaussian produces a Gaussian output, while a nonlinear system produces a non-Gaussian output. Hence Kalman filters work well with linear systems of equations. However, most of the real-time systems that are encountered are nonlinear. 

One way to apply Kalman filters to these nonlinear systems is to approximate them as linear systems by using Taylor series~\cite{Taylor} around the mean of the nonlinear Gaussian. This version of the Kalman filter is referred to as the Extended Kalman Filter (EKF). The EKF algorithm is as shown in Algorithm~\ref{EKF}. The terms used in the algorithm~\ref{EKF} are listed below.

\begin{table}[b]
\begin{tabular}[h]{l l l}
    ${x}$& - &State Matrix \\
    $\hat{x}$& - &State Prediction\\
    $f : X \times U \rightarrow R\textsuperscript{2}$& - &Function to predict the states \\
    $F$ & - &State Transition Matrix\\
    $\hat{P}$ & - &Predicted Covariance\\
\end{tabular}
\end{table}

\begin{table}[t]
\begin{tabular}{l l l}
    $P$ & - &Process Noise Covariance\\
    $Q$ & - &Measurement Noise Covariance\\
    $z$ & - &Measurement from Sensors\\
    $h(\hat{x})$ & - &Prediction to Measurement coordinate\\
    $y$ & - & \vtop{\hbox{\strut Difference between Prediction}\hbox{\strut and Measurement}}\\
    $H$ & - &Jacobian Matrix\\
    $R$ & - &Measurement Noise Covariance\\
    $K$ & - &Kalman Gain\\
    $I$ & - &Identity Matrix\\
    \textit{k} & - &Represents the current time step\\
    \textit{k-1} & - &Represents the previous time step\\
\end{tabular}
\end{table}

\begin{algorithm}[h]
\SetAlgoLined
\KwResult{State Matrix : \textit{x} }
\textbf{\textit{Initialization:}} 
Initialize the state of the filter(\textit{Q,P}) and the belief in the state(\textit{K})\\
 \While{New Datapoint}{
    \textbf{\textit{Predict:}}\\
    \hspace{10mm}$\hat{x}_{k}$ = \textit{f}(\textbf{x,u})\\
    \hspace{2mm}$F_{k} = \frac{\partial f(x,u)}{\partial x}|_{x,u}$\\
    \hspace{2mm}$\hat{P}_{k = F_{k}P_{k-1}F_{k}^{T} + Q}$
    
    \textbf{\textit{Update:}}\\
    $y_{k} = z_{k} - h(\hat{x}_{k})$\\
    \hspace{2mm}$H_{k} = \frac{\partial h(\hat{x})}{\partial x}|_{x}$\\
    \hspace{2mm}$S_{k} = H_{k}\hat{P}_{k}H^{T}_{k} + R$\\
    \hspace{2mm}$K_{k} = \hat{P}_{k}H^{T}_{k}S^{-1}_{k}$\\
    \hspace{2mm}$x_{k} = \hat{x}_{k} + K_{k}y_{k}$\\
    \hspace{2mm}$P_{k} = (I - K_{k}H_{k})\hat{P}_{k}$
 }
\caption{Extended Kalman Filter}
 \label{EKF}
\end{algorithm}

Applying the above mentioned Extended Kalman Filter Algorithm~\ref{EKF} to the RexROV submarine, the state matrix for 4 DoF shown in~\eqref{eq:35} can be found.

\begin{equation}
    \bm{x = }
    \begin{bmatrix}
    x\\
    y\\
    z\\
    \psi\\
    \end{bmatrix}
\label{eq:35}
\end{equation}

The states can be predicted using Newton's equations of motion\cite{Fossen_book} as shown in~\eqref{eq:36}. Here, \textit{v} is the velocity vector~\eqref{eq:37} and \textit{$\dot{v}$} is the acceleration vector~\eqref{eq:38}.

\begin{equation}
    \mathit{f}(x,u) = x_{k-1} + \mathit{v_{k}}\Delta t + \frac{\mathit{\dot{v}_{k}}\Delta t^{2}}{2}
\label{eq:36}
\end{equation}

\begin{equation}
    v = 
    \begin{bmatrix}
    u\\
    v\\
    w\\
    r\\
    \end{bmatrix}
\label{eq:37}
\end{equation}

\begin{equation}
    \dot{v} =
    \begin{bmatrix}
    \dot{u}\\
    \dot{v}\\
    \dot{w}\\
    \dot{r}\\
    \end{bmatrix}
\label{eq:38}
\end{equation}

The state transition matrix F in the expanded form is shown in ~\eqref{eq:39}.

\begin{equation}
F = 
    \begin{bmatrix}
    \frac{\partial f(x)}{\partial x} & \frac{\partial f(x)}{\partial y} & \frac{\partial f(x)}{\partial z} & \frac{\partial f(x)}{\partial \psi}\\
    \frac{\partial f(y)}{\partial x} & \frac{\partial f(y)}{\partial y} & \frac{\partial f(y)}{\partial z} & \frac{\partial f(y)}{\partial \psi}\\
    \frac{\partial f(z)}{\partial x} & \frac{\partial f(z)}{\partial y} & \frac{\partial f(z)}{\partial z} & \frac{\partial f(z)}{\partial \psi}\\
    \frac{\partial f(\psi)}{\partial x} & \frac{\partial f(\psi)}{\partial y} & \frac{\partial f(\psi)}{\partial z} & \frac{\partial f(\psi)}{\partial \psi}\\
    \end{bmatrix}
\label{eq:39}
\end{equation}

Using the previously discussed stability assumptions, the \textbf{F}-matrix reduces to~\eqref{eq:40}.
\begin{equation}
    F = 
    \begin{bmatrix}
    \frac{\partial f(x)}{\partial x} & 0 & 0 & 0\\
    0 & \frac{\partial f(y)}{\partial y} & 0 & 0\\
    0 & 0 & \frac{\partial f(z)}{\partial z} & 0\\
    0 & 0 & 0 & \frac{\partial f(\psi)}{\partial \psi}\\
    \end{bmatrix}
\label{eq:40}
\end{equation}

The terms in the F-matrix are derived as follows,
\begin{equation}
    \frac{\partial f(x)}{\partial x} = 1 + \frac{\partial u}{\partial x}\Delta t + \frac{\partial \dot{u}}{\partial x} \frac{\Delta t^2}{2}
\label{eq:41}
\end{equation}

\begin{equation}
    \frac{\partial f(x)}{\partial x} = 1 + \frac{\partial u}{\partial t}\frac{\partial t}{\partial x}\Delta t + \frac{\partial \dot{u}}{\partial t}\frac{\partial t}{\partial x} \frac{\Delta t^2}{2}
\label{eq:42}
\end{equation}

For small changes in $\dot{u}$ with respect to time t, $\frac{\partial \dot{u}}{\partial t}$ can be taken as $\frac{\Delta \dot{u}}{\Delta t}$ and~\eqref{eq:42} changes to,

\begin{equation}
    \frac{\partial f(x)}{\partial x} = 1 + \dot{u}\frac{1}{u}\Delta t + \frac{\Delta \dot{u}}{\Delta t}\frac{1}{u}\frac{\Delta t^2}{2}
\label{eq:43} 
\end{equation}

\begin{equation}
    \frac{\partial f(x)}{\partial x} = 1 + \frac{\dot{u}}{u}\Delta t + \frac{\Delta \dot{u}\Delta t}{2u}
\label{eq:44} 
\end{equation}

Similarly the other three terms in the \textbf{F}-matrix are obtained as shown in~\eqref{eq:45} to~\eqref{eq:47}.

\begin{equation}
    \frac{\partial f(y)}{\partial y} = 1 + \frac{\dot{v}}{v}\Delta t + \frac{\Delta \dot{v}\Delta t}{2v}
\label{eq:45} 
\end{equation}

\begin{equation}
    \frac{\partial f(z)}{\partial z} = 1 + \frac{\dot{w}}{w}\Delta t + \frac{\Delta \dot{w}\Delta t}{2w}
\label{eq:46} 
\end{equation}

\begin{equation}
    \frac{\partial f(\psi)}{\partial \psi} = 1 + \frac{\dot{r}}{r}\Delta t + \frac{\Delta \dot{r}\Delta t}{2r}
\label{eq:47} 
\end{equation}

In equations~\eqref{eq:44} to~\eqref{eq:47}, the linear velocity values are obtained from the DVL sensor and the angular velocity is obtained from the IMU sensor. The acceleration values are obtained from the dynamic equations~\eqref{eq:27} to~\eqref{eq:30}. This concludes the predict step in the EKF Algorithm~\ref{EKF}. The measurement matrix z is represented as shown in~\eqref{eq:48}.

\begin{equation}
    z = 
    \begin{bmatrix}
    x_{m}\\
    y_{m}\\
    z_{m}\\
    \psi_{m}\\
    \end{bmatrix}
\label{eq:48}
\end{equation}

Here, $x_{m}, y_{m}$ comes from the GPS sensor, $z_{m}$ comes from the pressure sensor and $\psi_{m}$ comes from the IMU. Also the prediction values are already in the same coordinates as the measurement values and hence h(\textit{$\hat{x}$}) is the same as $\hat{x}$. This results in the Jacobian matrix (H) becoming an Identity matrix ($I_4$) as shown in~\eqref{eq:49}.

\begin{equation}
    H = \frac{\partial h( \hat{x})}{\partial \hat{x}} = \frac{\partial\hat{x}}{\partial\hat{x}} = \mathit{I_4}
\label{eq:49}
\end{equation}

The other update equations are reduced to the following:

\begin{equation}
    S_{k} = H_{k}\hat{P}_{k} + R
\label{eq:50}
\end{equation}

\begin{equation}
    K_{k} = \hat{P}_{k}S^{-1}_{k}
\label{eq:51}
\end{equation}

\begin{equation}
    x_{k} = \hat{x}_{k} + K_{k}y_{k}
\label{eq:52}
\end{equation}

\begin{equation}
    P_{k} = (I - K_{k})\hat{P}_{k}
\label{eq:53}
\end{equation}

Thus, the EKF Algorithm for RexROV becomes as shown in Algorithm ~\eqref{EKF_Rexrov}. The algorithm has been applied to several test courses from~\cite{Norris} in the next section and their results are discussed. The source code for implementation of the EKF algorithm shown below is available in \url{https://gitlab.engr.illinois.edu/auvsl/submarine}

\begin{algorithm}[h]
\SetAlgoLined
\KwResult{State Matrix : \textbf{x} }
\textbf{\textit{Initialization:}} 
Initialize the state of the filter(\textbf{Q,P}) and the belief in the state(\textbf{K})\\
 \While{New Datapoint}{
    \textbf{\textit{Predict:}}\\
    \hspace{10mm}$\hat{x}_{k}$ = \textit{f}(x,u)\\
    \hspace{2mm}$F_{k} = \frac{\partial f(x,u)}{\partial x}|_{x,u}$\\
    \hspace{2mm}\textbf{$\hat{P}_{k} = F_{k}P_{k-1}F_{k}^{T} + Q$}
    
    \textbf{\textit{Update:}}\\
    $y_{k} = z_{k} - \hat{x}_{k}$\\
    \hspace{2mm}$S_{k} = H_{k}\hat{P}_{k} + R$\\
    \hspace{2mm}$K_{k} = \hat{P}_{k}S^{-1}_{k}$\\
    \hspace{2mm}$x_{k} = \hat{x}_{k} + K_{k}y_{k}$\\
    \hspace{2mm}$P_{k} = (I - K_{k})\hat{P}_{k}$
 }
\caption{Extended Kalman Filter for RexROV}
 \label{EKF_Rexrov}
\end{algorithm}

In the dynamic Kalman filter, the velocity values are taken from the Doppler Velocity Logs (DVL) and thrust values are obtained from the pre-generated look-up table. Using these, the $\hat{x}_{k}$ values were calculated using the dynamic equations mentioned previously. For the kinematic filter, velocity values from DVL and acceleration values from the IMU are used in Newton's equations of motion to calculate $\hat{x}_{k}$. The update step remains the same for both the filters, with the GPS providing \textbf{x} and \textbf{y} co-ordinates, the pressure sensor providing the \textbf{z} co-ordinate and the compass providing the heading $\psi$ for the jacobian matrix.
\section{Pure Pursuit Controller}
 For a given set of waypoints, a pure pursuit algorithm approaches the target waypoint ignoring the dynamics/kinematics of the environment and the vehicle (except vehicle length). Once in the vicinity (user defined) of the target waypoint, the algorithm generates the steering command for the next waypoint until it reaches the last waypoint~\cite{PurePursuit}. For the purposes of this work, the steering command is generated while the linear velocity is constant. \\ 
\begin{figure}[h]
\centering
\includegraphics[scale=0.65]{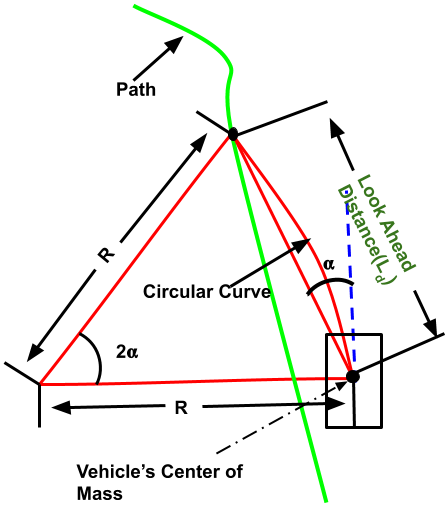}
\caption{Pure Pursuit Model}
    \label{fig:purepursuit}
\centering
\end{figure}

Considering RexROV as a point mass~\cite{thirteen}, Pure Pursuit aims to align the Center of Mass (in our case) along the reference trajectory through generating steering commands using the following formulations.
\begin{equation}
    K = \frac{2*sin(\alpha)}{L_{d}}\\
\label{eq:54}
\end{equation}

\begin{equation}
    \delta = tan^{-1}(K*L)
\label{eq:55}
\end{equation}

The vehicle length (${L}$) and the look ahead distance ($L_{d}$) are taken as unity as well. ${\alpha}$ is the difference between vehicle pose and the slope of the line connecting the target point and the last waypoint. $x_k$ in the EKF algorithm~\ref{EKF_Rexrov} gives the current pose of the RexROV. Both slope and pose are converted within a 0 to 2${\pi}$ range.
Output ${\delta}$ is published to the $cmdvel$ Rostopic with a gain of 0.3.

\section{Experiments and Results}
Experiments were conducted on open source simulation software Gazebo 7. For underwater scenario and sensors simulation, UUVSIM \cite{uuvsim} is used in parallel with ROS Kinetic.\par 
The vehicle, RexROV, used is $2.6\times 1.5\times 1.6$ $m^3$ in dimensions and 1863 kg in weight. Max speed of vehicle is 0.3 m/s and only steering is controlled using  pure pursuit controller. Vehicle is running under a depth of 20 m from sea level. Wind and current disturbance is considered zero for the purpose of demonstration i.e. no external disturbance. \par
In this section, the performance of the kinematic and the dynamic model driven pure pursuit controllers are compared on various test courses adopted from~\cite{Norris}.For comparison, the ground truth localization is obtained by subscribing to topic \textbf{$/rexrov/pose\textunderscore gt$} which is published by default when vehicle is launched with required inertial navigation sensors. However, it is not mentioned \cite{uuvsim} how these sensor readings are processed to obtain ground truth value.   \par
The figure shows the actual path followed by the vehicle as opposed to the output of the dynamic model and kinematic model. The results from the test courses are summarized as a table. \textbf{\textit{Total}} is the mean of the Cartesian distance between the \textit{reference path} and \textit{the path taken by the RexROV} when using the controllers. \textbf{\textit{X-Kalman}}  (along x axis) gives the root mean squared error (RMSE) between the value predicted by the \textit{Kalman filter} and \textit{the actual path} traced by the RexROV. Likewise, \textbf{\textit{Y-Kalman}} (along y axis) gives MSE in the y-axis.
 \subsection{Behavior Evaluation 1 }

 \begin{figure}[h]
\centering
\includegraphics[scale=0.60]{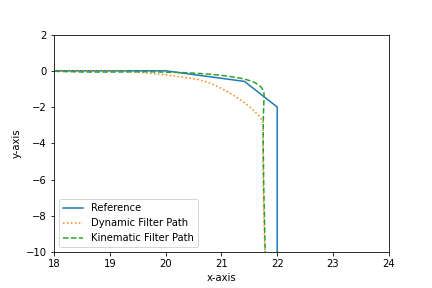}
\caption{Behavior Evaluation 1}
    \label{fig:bp1}
\centering
\end{figure}

The first test course involves moving along a straight line followed by a fixed turning radius and a sharp turn. As seen in Fig.~\ref{fig:bp1}, the kinematic filter follows the reference path much closer than the dynamic filter. After the simple turn, both the filters converge at the same rate.

\begin{table}[h]
    \caption{Results of Behavior Evaluation 1}
    \centering
    \begin{tabular}{|c|c|c|}
    \hline
    Error&Dynamic Filter&Kinematic Filter\\
    \hline
    Total&0.1166m&0.09880m\\
    \hline
    X-Kalman&0.0825m&0.3432m\\
    \hline
    Y-Kalman&0.2665m&1.024m\\
    \hline
    \end{tabular}
    \label{table:ResultsBP1}
\end{table}

From Table~\ref{table:ResultsBP1}, the total error of the kinematic filter is less than the dynamic filter. However, this is attributed to the controller and not the filter itself as it can be inferred from the other two entries in the table. To make further assessment of the filters, they were tested on the behavior evaluation 2 test course.
 
 \subsection{Behavior Evaluation 2 }
The second test course is a sinusoidal path which uses curves of constant radius with both clockwise and counter-clockwise directions. This test course helps in determining if there are handling and navigation issues.

\begin{figure}[h]
\centering
\includegraphics[scale=0.60]{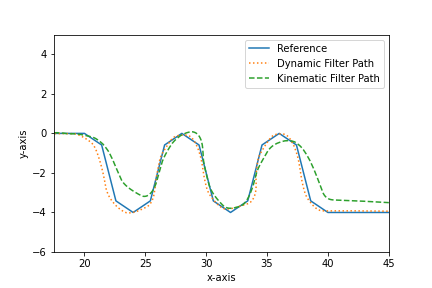}
\caption{Behavior Evaluation 2}
    \label{fig:bp2}
\centering
\end{figure}

\vspace{0.6cm}

As seen in Fig.~\ref{fig:bp2},the dynamic filter based localization estimates are  much closer to the reference trajectory as opposed to kinematic approach based. Dynamic filter looks stable while kinematic filter looks diverging first and then converging again. This behaviour can be attributed to the acceleration values obtained from the IMU. In Fig.~\ref{fig:imu-x}, IMU acceleration is leading or lagging through out with respect to Dynamic acceleration, hence such fluctuations in kinematic filter is observed. 
 
 \begin{table}[h]
    \caption{Results of Behavior Evaluation 2}
    \centering
    \begin{tabular}{|c|c|c|}
    \hline
    Error&Dynamic Filter&Kinematic Filter\\
    \hline
    Total& 0.1659m&0.2482m\\
    \hline
    X-Kalman&0.1213m&0.39145m\\
    \hline
    Y-Kalman&0.3682m&4.336m\\
    \hline
    \end{tabular}
    \label{table:ResultsBP2}
\end{table}
\vspace{0.5cm}
From table~\ref{table:ResultsBP2}, the mean error of the dynamic filter is less than the kinematic filter. The dynamic filter is able to handle the simultaneous changes in the \textit{x}-axis and \textit{y}-axis much better than the kinematic filter.
 \subsection{Behavior Evaluation 3}
The third test course is a U-shaped path with clockwise and counter-clockwise curves and straightaways.
\begin{figure}[h]
\centering
\includegraphics[scale=0.60]{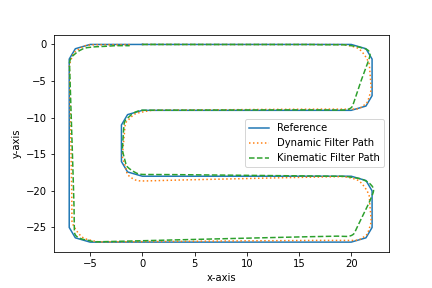}
\caption{Behavior Evaluation 3}
    \label{fig:bp3}
\centering
\end{figure}

 \begin{table}[h]
    \caption{Results of Behavior Evaluation 3}
    \centering
    \begin{tabular}{|c|c|c|}
    \hline
    Error&Dynamic Filter&Kinematic Filter\\
    \hline
    Total&0.0738m&0.1965m\\
    \hline
    X-Kalman&0.2462&0.4028\\
    \hline
    Y-Kalman&0.4003m&4.0014m\\
    \hline
    \end{tabular}
    \label{table:ResultsBP3}
\end{table}

From Fig.~\ref{fig:bp3}, in behavior evaluation course 3, which includes all possible movements of the ROV, the dynamic filter converges to the path more quickly than its kinematic counter part. Again the slight divergence on the right hand side is observed in kinematic filter. The reason is similar to that of given while evaluating BE2, i.e., acceleration from the IMU is fluctuating at these points.

\section{Practical Implementation}
In order to validate the results found during the simulated experiments, physical experiments will be needed. While beyond the scope of the present paper, this section provides some guidelines for accomplishing the physical validation. 

\begin{enumerate}
    \item First, the work presented in this paper is dependent on three major assumptions: 
    \begin{enumerate}
        \item That a valid dynamic model can be created which is accurate enough to predict the behavior of the system.
        \item That a valid kinematic model can be created of the system which can also predict the behavior of the system and that it generally had poorer performance with the Kalman filters than a dynamic model would be.
        \item That the reduced-order dynamic model with four 4 DoF is rigorous enough to capture the behavor of the system appropriately.
    \end{enumerate}
    \item The first and most important thing to do is to experimentally validate both the dynamic and kinematic models of the system (which could be partially done by reviewing the literature in depth). If the differences between them are inconclusive for a wide range of scenarios, the one which provides the lowest computational cost should be used.
    \item The assumption that a six DoF model and four DoF model are equivalent for this type of under-water vehicle needs experimental validation. 
    \item Once the assumptions are validated (or at least the ranges under which they are valid are identified), detailed sensitivity functions for each (and other parts of the system) can be calculated which will help identify areas of risk in the system. 
    \item From here, the validated models, localization, and control systems can be used as the basis for new vehicle designs approaches, both for plant design and controller design (or both at the same time, as done in co-design~\cite{Chilan2017, Allison2014}).
\end{enumerate}

\section{Summary, Conclusions, and Future Work}

In this research, the success of a reduced order dynamic model with an Extended Kalman Filter was demonstrated in simulation. Localization values obtained were very close to ground truth and no large deviations were observed. Comparison with Kinematics based Kalman filter justified the choice of including dynamics in the form of a reduced order model. Dynamics-based EKF performance surpassed its counterpart in all test cases, except one, with large margins.

From the Figs.~\ref{fig:bp1} to~\ref{fig:bp3} and the Tables~\ref{table:ResultsBP1} to~\ref{table:ResultsBP3}, the dynamic Kalman filter predictions were very close to the ground truth values. On comparison of the individual values predicted by the dynamic filter and the kinematic filter with the actual position of RexROV, the dynamic filter was more stable and reliable than the kinematic filter. The results were based on movement only in the two primary axes, \textit{x} and \textit{y}, similar to a ground vehicle. Further research is being conducted on the simultaneous triaxial movement to make the controller universal. It is expected that the movement on the \textit{z}-axis should not affect the performance of the dynamic Kalman filter. Currently the model uses GPS data to correct position estimation. However research has indicated that the strength of the electromagnetic waves for the GPS signal reduces significantly underwater~\cite{gpsprob}. Methods such as station keeping, SONARSLAM and vision systems are currently being explored as an alternative for position estimation. Alternative controller techniques are being researched to improve the performance of the model. The predictions were conducted at a frequency of 10Hz  and the dynamic filter worked on the simulation without any performance issues. However, the filters need to be implemented in real-time hardware to assess the actual computational performance.

\begin{acknowledgment}
This research was funded by a generous gift grant from TechnipFMC (\url{https://www.technipfmc.com/en}).
\end{acknowledgment}

%




\end{document}